\documentclass[letterpaper, 10 pt, conference]{ieeeconf}  

\IEEEoverridecommandlockouts                              

\usepackage{acro}
\usepackage{graphicx}

\usepackage{amsmath}
\usepackage{amssymb}
\usepackage{bm}
\usepackage{siunitx}
\usepackage{makecell}
\usepackage{multirow}
\usepackage{multicol}
\usepackage{caption} 
\usepackage{subcaption}
\usepackage{subcaption}
\usepackage{booktabs} 
\usepackage{tikz}
\usetikzlibrary{decorations.pathreplacing}
\usetikzlibrary{shapes.geometric, arrows, positioning, calc, decorations.pathreplacing}
\usepackage{hyperref}

\usepackage{graphicx}   
\usepackage{tabularx}   
\usepackage{array}
\usepackage{xcolor}     
\usepackage[table]{xcolor}
\usepackage{pgfplots}
\usepackage{pgfplotstable}
\pgfplotsset{compat=1.18}
\usepackage{filecontents}

\usepackage[symbol]{footmisc}

\graphicspath{ {./images/} }
\newcommand{\eg}{e.g., }

\usepackage{textcomp}
\newcommand\copyrighttext{%
	\footnotesize \textcopyright 2025 IEEE. Personal use of this material is permitted. Permission from IEEE must be obtained for all other uses, in any current or future media, including reprinting/republishing this material for advertising or promotional purposes, creating new collective works, for resale or redistribution to servers or lists, or reuse of any copyrighted component of this work in other works. DOI: \hyperlink{10.1109/IV64158.2025.11097767}{10.1109/IV64158.2025.11097767}.}
\newcommand\copyrightnotice{%
	\begin{tikzpicture}[remember picture,overlay]
	\node[anchor=south,yshift=10pt] at (current page.south) {\fbox{\parbox{\dimexpr\textwidth-\fboxsep-\fboxrule\relax}{\copyrighttext}}};
	\end{tikzpicture}%
}

\title{\LARGE \bf

Assessing the Completeness of Traffic Scenario Categories for Automated Highway Driving Functions via Cluster-based Analysis

}

\author{Niklas Roßberg$^{1}$, Marion Neumeier$^{1}$, Sinan Hasirlioglu$^{2}$, Mohamed Essayed Bouzouraa$^{2}$ and Michael Botsch$^{1}$
\thanks{$^{1}$Technische Hochschule Ingolstadt, 85049 Ingolstadt, Germany
        {\tt\small firstname.lastname@thi.de}}%
\thanks{$^{2}$AUDI AG, 85045 Ingolstadt, Germany
        {\tt\small \{sinan.hasirlioglu, essayed.bouzouraa\}@audi.de}}
}

\DeclareAcronym{ADS}{
short = ADS ,
long = Automated Driving Systems ,
short-plural = s ,
long-plural = s 
}

\DeclareAcronym{CCP}{
short = CCP ,
long = Coupon Collector's Problem ,
short-plural = s ,
long-plural = s 
}

\DeclareAcronym{VQ}{
short = VQ,
long = Vector Quantization,
short-plural = s ,
long-plural = s 
}

\DeclareAcronym{CVQ-VAE}{
short = CVQ-VAE ,
long = Clustering Vector Quantized - Variational Autoencoder ,
short-plural = s ,
long-plural = s 
}

\DeclareAcronym{kl}{
short = kl ,
long = keep lane ,
short-plural = s ,
long-plural = s 
}

\DeclareAcronym{lcl}{
short = lcl ,
long = lane change left ,
short-plural = s ,
long-plural = s 
}

\DeclareAcronym{lcr}{
short = lcr ,
long = lane change right ,
short-plural = s ,
long-plural = s 
}

\DeclareAcronym{ML}{
short = ML ,
long = Machine Learning ,
short-plural = s ,
long-plural = s 
}

\DeclareAcronym{ODD}{
short = ODD ,
long = Operational Design Domain ,
short-plural = s ,
long-plural = s 
}

\DeclareAcronym{OOD}{
short = OOD ,
long = Out-of-Distribution ,
short-plural = s ,
long-plural = s 
}

\DeclareAcronym{PDF}{
short = PDF ,
long = probability density function ,
short-plural = s ,
long-plural = s 
}

\DeclareAcronym{VAE}{
short = VAE ,
long = Variational Autoencoder ,
short-plural = s ,
long-plural = s 
}

\DeclareAcronym{VQ-VAE}{
short = VQ-VAE ,
long = Vector Quantized - Variational Autoencoder ,
short-plural = s ,
long-plural = s 
}

\begin{document}

\maketitle
\copyrightnotice
\thispagestyle{empty}
\pagestyle{empty}

\begin{abstract}

The ability to operate safely in increasingly complex traffic scenarios is a fundamental requirement for \ac{ADS}.
Ensuring the safe release of \ac{ADS} functions necessitates a precise understanding of the occurring traffic scenarios.
To support this objective, this work introduces a pipeline for traffic scenario clustering and the analysis of scenario category completeness.
The \ac{CVQ-VAE} is employed for the clustering of highway traffic scenarios and utilized to create various catalogs with differing numbers of traffic scenario categories. 
Subsequently, the impact of the number of categories on the completeness considerations of the traffic scenario categories is analyzed. 
The results show an outperforming clustering performance compared to previous work.
The trade-off between cluster quality and the amount of required data to maintain completeness is discussed based on the publicly available highD dataset.

\end{abstract}
\vspace{-0pt}

\section{Introduction}
A key challenge for the safe release of automated driving is equipping Automated Driving System \mbox{(\ac{ADS})} functions with the ability to accurately interpret and respond to the wide range of driving conditions they encounter.
To ensure the safety of the system and maintain compliance with legal and regulatory requirements, this challenge is typically addressed through the concept of an \ac{ODD}.
It specifies the conditions under which the designed system has to operate safely. 
Precisely defining the~\ac{ODD} and thoroughly understanding the traffic scenarios occurring within it are essential for the reliable execution of critical downstream tasks.
These include behavior prediction~\cite{GALVAO2024121983, POMDPTIM, NEIMEIERGFFT} and motion planning~\mbox{\cite{ReviewMOTIONPL2023, mao2023gptdriver}}. 
A widely used approach to help understanding the occurring traffic scenarios is to categorize them.
This approach assumes that the theoretically infinite number of possible scenarios can be grouped into a finite set of categories~\cite{Kruber2019}.
One major advantage of this method is the ability to determine whether a new scenario is similar to an already known category and can thus be considered familiar.
This can, for instance, enhance the accuracy of future trajectory predictions~\cite{Neumeier2024}.
Another key benefit of a successful categorization is that it enables an assessment of data sufficiency. 
Specifically, it helps determine whether the identified scenario categories capture the full diversity and range of scenarios collected for the corresponding \ac{ODD}~\cite{Hauer2019}.
In other words, it enables an evaluation of whether the scenario category dataset can be considered complete.

The most common approaches for traffic scenario categorization include rule-based strategies and \mbox{\ac{ML}} approaches. 
As capturing the complexity of real-world scenarios with rule-based algorithms proves challenging, there is an increasing focus on researching \ac{ML} approaches, \eg \cite{Neumeier2024, CHETOUANE2022100377, bekkair2023GAT, HARMENING2020}. 
\begin{figure}[]
    \centering
    \begin{tikzpicture}[node distance= 2cm, scale=0.65, , every node/.style={transform shape}]
        \tikzstyle{decision} = [diamond, draw, text centered, inner sep=5pt, line width = 1.3pt, font=\scriptsize]

        \node (A) [draw, rectangle, minimum width=1.5cm, minimum height=1.0cm, align=center, rounded corners = 2.4, line width = 0.5pt, font=\small] at (-9.5,0) {Data\\Collection};

        \node (B) [above of=A, node distance= 2.2cm, draw, rectangle, minimum width=1.5cm, minimum height=1.0cm, align=center, line width = 0.5pt, font=\small] {};

        \node (B2) [node distance= 0cm, fill=white, draw, rectangle, minimum width=1.5cm, minimum height=1.0cm, align=center, line width = 0.5pt, font=\scriptsize] at ($(B.south west) + (0.7,0.45)$) {};  
        \node (B3) [node distance= 0cm, fill=white, draw, rectangle, minimum width=1.5cm, minimum height=1.0cm, align=center, line width = 0.5pt, font=\small] at ($(B2.south west) + (0.7,0.45)$) {Driving \\Data};

        \node (C) [right of=B, draw, rectangle, minimum width=1.5cm, minimum height=1.0cm, align=center, rounded corners = 2.4, line width = 0.5pt, font=\small] {Automatic\\Clustering};

        \node (D) [right of=C, node distance= 2.2cm, draw, rectangle, minimum width=1.5cm, minimum height=1.0cm, align=center, line width = 0.5pt, font=\scriptsize] {};

        \node (D2) [node distance= 0cm, fill=white, draw, rectangle, minimum width=1.5cm, minimum height=1.0cm, align=center, line width = 0.5pt, font=\scriptsize] at ($(D.south west) + (0.7,0.45)$) {};
        \node (D3) [node distance= 0cm, fill=white, draw, rectangle, minimum width=1.5cm, minimum height=1.0cm, align=center, line width = 0.5pt, font=\small] at ($(D2.south west) + (0.7,0.45)$) {Scenario\\Categories};
        \node (E) [right of=D, node distance= 2.2cm, draw, rectangle, minimum width=1.5cm, minimum height=1.0cm, align=center, rounded corners = 2.4, line width = 0.5pt, font=\small] {Completeness\\Check};

        \node (F) [decision, right of=E,] {};
        \node [above = 0.0cm of F, font=\small] {Complete?};
        \node [below right = -0.1 cm and 0.0cm of F, font=\small] {Yes};
        \node [below left = 0.4 cm and -0.2cm of F, font=\small] {No};
        \node (G) [right of=F, node distance= 1.8cm,draw, rectangle, minimum width=1.5cm, minimum height=1.0cm, align=center, rounded corners = 2.4, line width = 0.5pt, font=\small] {Test Case\\Generation};

        \draw[->, line width=1.0pt] (A) -- ++(0,1.6);
        \draw[->, line width=1.0pt] (B) -- (C);
        \draw[->, line width=1.0pt] (C) -- ++(1.3,0);
        \draw[->, line width=1.0pt] (D) -- (E);
        \draw[->, line width=1.0pt] (E) -- (F);
        \draw[->, line width=1.0pt] (F) -- (G);
        \draw[->,line width=1.0pt] (F.south) |- (A.east);
        \draw[->, line width=1.0pt] (B.north) -- ++(0,0.5) -| (E.north);

        \node (K) [draw, rectangle, minimum width=11.5cm, minimum height=2.7cm, rounded corners = 4.4, line width = 1.0pt, fill=lightgray!15, font=\small] at (-4.5,-2.5) {}; 
        \node [below left = -2.55cm and 0.6cm of K, font=\large, rotate = 90] {CVQ-VAE};

        \node (KW1) [left of=K, node distance= 5.0cm, draw, rectangle, minimum width=1.0cm, minimum height=1.0cm, align=center, line width = 0.5pt, font=\small] {};
        \node (KW2) [node distance= 0cm, fill=white, draw, rectangle, minimum width=1.0cm, minimum height=1.0cm, align=center, line width = 0.5pt, font=\scriptsize] at ($(KW1.south west) + (0.45,0.45)$) {};  
        \node (KW3) [node distance= 0cm, fill=white, draw, rectangle, minimum width=1.0cm, minimum height=1.0cm, align=center, line width = 0.5pt, font=\small] at ($(KW2.south west) + (0.45,0.45)$) {};

        \node (KW4) [left of=K, node distance= -5.1cm, draw, rectangle, minimum width=1.0cm, minimum height=1.0cm, align=center, line width = 0.5pt, font=\small] {};
        \node (KW5) [node distance= 0cm, fill=white, draw, rectangle, minimum width=1.0cm, minimum height=1.0cm, align=center, line width = 0.5pt, font=\scriptsize] at ($(KW4.south west) + (0.45,0.45)$) {};  
        \node (KW6) [node distance= 0cm, fill=white, draw, rectangle, minimum width=1.0cm, minimum height=1.0cm, align=center, line width = 0.5pt, font=\small] at ($(KW5.south west) + (0.45,0.45)$) {};
        
        \node (KE) [draw, trapezium, trapezium left angle=60, trapezium right angle=60, minimum width=2.2cm, minimum height=0.75cm, fill=cyan!10, line width = 1.0pt, rotate =-90, text centered, font=\large] at (-8.35,-2.5) {}; 
        \node [below right = -0.85 cm and 0.25cm of KE, font=\large] {E};
        \node [below right = -0.7 cm and -1.0cm of KE, font=\large] {$\bm{\xi}$};
        \node (KE2) [right of=KE,node distance= 1.0cm, draw, rectangle, minimum width=1.0cm, minimum height=1.0cm, fill=cyan!10, align=center, line width = 1.0pt, font=\Large]  {$\hat{\bm{z}}$};

        \node (KE3) [right of=KE,node distance= 6.7cm, draw, rectangle, minimum width=1.0cm, minimum height=1.0cm, fill=cyan!10, align=center, line width = 1.0pt, font=\Large]  {$\bm{z}_{q}$};
        \node (KD) [right of=KE3, node distance= 1.0cm, draw, trapezium, trapezium left angle=60, trapezium right angle=60, minimum width=2.2cm, minimum height=0.75cm, fill=cyan!10, line width = 1.0pt, rotate =90, text centered, font=\scriptsize] {};
        \node [below left = -0.85 cm and 0.25cm of KD, font=\large] {D};
        \node [below left = -0.8 cm and -0.85cm of KD, font=\large] {$\bm{\hat{\xi}}$};

        \draw [decorate,decoration={brace,amplitude=6pt, raise=8pt},  line width=1.0pt, dashed] (-9.0,-1.5) -- (-6.0,-1.5) node [black,midway,xshift=-0.6cm] {};
        \draw [decorate,decoration={brace,amplitude=6pt,raise=8pt},line width=1.0pt, dashed] (-2.8,-1.5) -- (0.2,-1.5) node [black,midway,xshift=-0.6cm] {};

        \node (QB) [draw, rectangle, minimum width=3cm, fill=white, minimum height=2.5cm, align=center, rounded corners = 3.4, line width = 1.0pt, font=\scriptsize] at (-4.5,-2.5) {};
        \node [above = -0.55cm of QB, font=\footnotesize] {Codebook $\mathit{Q}$ };

        \node (QB2) [above right = -1cm and -2.9cm of QB, draw, rectangle, minimum width=0.41cm, minimum height=0.41cm, fill=white, align=center, line width = 0.7pt, font=\footnotesize] {1};
        \node (QB21) [right = 0.0cm of QB2,draw, rectangle, minimum width=2.3cm, minimum height=0.41cm, fill=lime!15, align=center, line width = 0.7pt, font=\footnotesize]  {$\bm{z}_{1} \quad \rightarrow \quad  p_1$};
        \node (QB3) [above right = -0.87cm and -0.43cm of QB2, draw, rectangle, minimum width=0.41cm, minimum height=0.41cm, fill=white, align=center, line width = 0.7pt, font=\footnotesize]{2};
        \node (QB31) [right = 0.0cm of QB3,draw, rectangle, minimum width=2.3cm, minimum height=0.41cm, fill=lime!40, align=center, line width = 0.7pt, font=\footnotesize]  {$\bm{z}_{2} \quad \rightarrow \quad p_2$};
        \node (QB4) [above right = -0.87cm and -0.43cm of QB3, draw, rectangle, minimum width=0.41cm, minimum height=0.41cm, fill=white, align=center, line width = 0.7pt, font=\footnotesize] {3};
        \node (QB41) [right = 0.0cm of QB4,draw, rectangle, minimum width=2.3cm, minimum height=0.41cm, fill=lime, align=center, line width = 0.7pt, font=\footnotesize]  {$\bm{z}_{3} \quad \rightarrow  \quad p_3$};
        \node (QB5) [above right = -1.07cm and -0.43cm of QB4, draw, rectangle, minimum width=0.41cm, minimum height=0.41cm, fill=white, align=center, line width = 0.7pt, font=\footnotesize] {$Q$};
        \node (QB51) [right = 0.0cm of QB5,draw, rectangle, minimum width=2.3cm, minimum height=0.41cm, fill=lime!55, align=center, line width = 0.7pt, font=\footnotesize]  {$\bm{z}_{Q} \quad \rightarrow \quad  p_Q$};
        % ...
        \node [below right = -0.08cm and 0.9cm of QB4, font=\large] {... };

        \draw[->, line width=1.0pt] (KE2) -- (QB);
        \draw[->, line width=1.0pt] (QB) -- (KE3);

        \draw[<-, dashed,  line width=1.0pt] (C.south) -- (-7.48,-0.5);
        \draw[<-, dashed, line width=1.0pt] (-5.5, 1.6) -- (-5.5, -0.5);
        \draw[-, dashed, line width=1.0pt] (-5.5, -0.5) -- (-5.0, -0.5);
        \draw[-, dashed, line width=1.0pt] (-5.0, -0.5) -- (-5.0, -1.0);

        \draw[-, dashed, line width=1.0pt] (-3.6, -1.0) -- (-3.6, -0.5);
        \draw[-, dashed, line width=1.0pt] (-3.6, -0.5) -- (-3.1, -0.5);
        \draw[->, dashed, line width=1.0pt] (-3.1, -0.5) -- (-3.1, 1.7);

        \draw[-, dashed, line width=1.0pt] (-1.3, -0.7) -- (-1.3, -0.5);
        \draw[-, dashed, line width=1.0pt] (-1.3, -0.5) -- (0.7, -0.5);
        \draw[->, dashed, line width=1.0pt] (0.7, -0.5) -- (0.7, 1.7);

    \end{tikzpicture}
    
    \caption{Pipeline for identifying scenario categories and evaluating data completeness based on \cite{Neumeier2024, Hauer2019}.}
    \vspace{-1.5em}
    \label{fig:CVQ-VAE_CCP_Concept}
\end{figure}
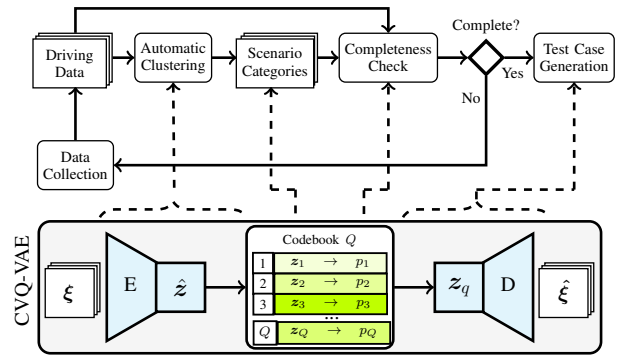
This work presents a \ac{ML}-based pipeline for clustering traffic scenarios and determining the amount of data required for a scenario category dataset to be considered complete, depending on the number of predefined scenario categories.
In Fig.~\ref{fig:CVQ-VAE_CCP_Concept} the proposed pipeline is visualized.
The first step, the scenario clustering, is based on~\cite{Neumeier2024}, which utilized the \ac{VQ-VAE}~\cite{Oord2017} for this task.
Thereby, the traffic scenarios are assigned to individual entries within a traffic scenario catalog.
This catalog is referred to as codebook. Its individual entries are termed codebook entries to stay consistent with literature~\cite{Oord2017,Zheng2023}. 
The \mbox{\ac{VQ-VAE}} employs \ac{VQ} to discretize continuous latent representations. 
Despite the effectiveness of \mbox{\ac{VQ-VAE}}  for traffic scenario clustering, codebook collapse, i.e., the under-utilization of the available codebook entries~\cite{Zheng2023}, limits its performance~\cite{Neumeier2024}.
This motivates the exploration of the \mbox{\ac{CVQ-VAE}}~\cite{Zheng2023} in this study.
By incorporating a dynamically initialized codebook, the \mbox{\ac{CVQ-VAE}} mitigates the issue of codebook collapse and enhances the utilization of codebook entries throughout training.
Utilizing this architecture, multiple codebooks are created, each comprising a different number of scenario categories.
The clustering results of this study are compared to those presented in~\cite{Neumeier2024}. 
In the second step, the impact of the number of scenario categories on the amount of data required to achieve a status of completeness is analyzed.
Given the theoretically infinite number of possible real-world scenarios, defining a test-ending criterion is essential.
To establish such a criterion, the~\ac{CCP}~\cite{Hauer2019} is applied.

The key contributions of this paper include:
\begin{enumerate}
\item Introduction of a novel pipeline to (a) identify traffic scenario categories with an adapted \ac{CVQ-VAE} and (b) evaluate the data completeness. 
\item Investigation on how the number of traffic scenario categories affects the performance of clustering and the scenario category completeness.
\item Evaluation of the pipeline on the publicly available highD dataset \cite{Krajewski2018}.

\end{enumerate}


\section{Related Work}

\subsection{Categorization and clustering of traffic scenarios}
Rule-based methods are among the most common and intuitive approaches for the categorization of traffic scenarios. 
The aim of these methods is to categorize basic traffic scenarios through the introduction of \eg feature thresholds and parameter ranges using expert knowledge \cite{Montanari2021, Montanari2021a, Tenbrock2021, Gelder2024}. 
Although rule-based methods can accurately identify scenarios, they depend strongly on expert knowledge. 
In addition, these approaches may fail to generalize the categorization of traffic scenarios, as the complex, variable, and context-dependent nature of traffic conditions makes it difficult to define exhaustive rules.
To complement rule-based methods, \ac{ML}-methods are increasingly used to  cluster traffic scenarios. 
A variety of unsupervised approaches have been applied to cluster traffic scenarios, including graph-based methods \cite{bekkair2023GAT,JU2023359}, contrastive learning frameworks \cite{Finamore2023}, and autoencoder-driven techniques \cite{Neumeier2024, LI2023110176, Balasubramanian2023}.  
While graph-based methods can suffer from unstable training characteristics \cite{NeumeierGAT}, and contrastive learning frameworks exhibit slow convergence \cite{osti_10321683}, autoencoder-driven techniques stand out due to their robust training behavior and ability to effectively handle complex data structures. 
Due to these advantages, autoencoders are widely utilized in various clustering-related works.
The \ac{VQ-VAE}, a method rooted in the \ac{VAE} \cite{Kingma2013} framework, has recently gained notable attention  \cite{Neumeier2024, Oord2017, Balasubramanian2023,  Idoko2024,  Tjandra2019}. 
The \ac{VQ-VAE} enhances the standard autoencoder by introducing a discrete latent space with vector quantization, allowing it to learn representations that are more compact and suitable for tasks like generative modeling.
Balasubramanian et al. \cite{Balasubramanian2023} utilize the \ac{VQ-VAE} within a open-world learning framework for traffic scenarios. 
Tjandra et al. \cite{Tjandra2019} applied the \ac{VQ-VAE} to discover subword units and generate compressed speech representations for synthesizing speech in a target voice.
The research most closely related to this work is that of  Neumeier et al. \cite{Neumeier2024}.
There the authors use the \ac{VQ-VAE} in the context of trajectory prediction with the goal to cluster traffic scenarios into a finite set of scenario representatives. 
However, it has been shown that the \ac{VQ-VAE} fails to fully utilize the entire codebook for the scenario clustering, potentially contributing to its limited performance. 
As discussed in \cite{Zheng2023}, this phenomenon is a common problem.
Consequently, exploring strategies to optimize codebook usage to refine scenario clustering opens up an area requiring further research efforts.

\subsection{Completeness of traffic scenario clusters}

Determining the completeness of traffic scenarios for \ac{ADS} validation is challenging as the theoretically infinite variability of scenarios within an \ac{ODD} makes it impossible to cover all possible scenarios within a dataset. 
In literature there are various approaches addressing the completeness of traffic scenarios \cite{Hauer2019, Gelder2024,Gelder2019,  Glasmacher2024}. 
For example, Gelder et al. in \cite{Gelder2024} and \cite{Gelder2019} use tags to describe traffic scenarios.  Based on those tags, a \ac{PDF} estimation is employed to assess how effectively an \ac{ODD} is covered. 
However, this approach has inherent limitations: it relies heavily on expert knowledge and cannot detect previously unknown scenarios that are not covered by the predefined tags. 
Other studies address the completeness of scenario concepts by exploring alternative methodologies, such as logic-based classifiers \cite{Glasmacher2024, schallau2023treebased}.
Hauer et al. \cite{Hauer2019} leverage the \ac{CCP} to determine the certainty with which a dataset can be considered complete by utilizing traffic scenario categories and their respective probabilities of occurrence. 
They conceptualize this as an urn problem.
A new scenario category with a predefined probability is introduced.
Subsequently, the total number of individual scenarios required in a dataset to ensure, with a given level of certainty, that all scenario categories are covered can be computed.
If this condition is met, the dataset is considered complete.
Although this approach is not entirely free of expert knowledge, it requires significantly less expertise compared to the methods described above.

\section{Preliminaries}

This study employs the \ac{CVQ-VAE} in order to cluster traffic scenarios and utilizes its outcomes for the \ac{CCP} to evaluate the completeness of scenario categories.
The concepts essential for understanding this approach are discussed in detail in this section. 
These include the dynamically initialized codebook of the \mbox{\ac{CVQ-VAE}} and the mathematical theory underlying the~\ac{CCP}.

\subsection{The updating mechanism of the \ac{CVQ-VAE}}
\label{subsec:CVQ-VAE}

The \ac{CVQ-VAE} builds upon the the \ac{VQ-VAE} \cite{Oord2017}, an autoencoder architecture derived from the classic \ac{VAE} \cite{Kingma2013}. 
In comparison to the \ac{VAE}, the \ac{VQ-VAE} introduces a codebook to quantize latent representations before reconstruction.
The \ac{VQ-VAE} is effective for representing discrete latent variables but suffers from codebook collapse. To address this, the authors of \cite{Zheng2023} introduced the \ac{CVQ-VAE}.
The \ac{CVQ-VAE} enhances the usage of the codebook entries through a dynamically-initialized vector quantized codebook~\cite{Zheng2023}. 
The core idea is to update the codebook entries $q$ depending on their average usage. 
To achieve this, the average usage in the $(t)$-th minibatch $N_{q}^{(t)}$ for each codebook entry $\bm{z}_q$ is initialized as $N_q^{(0)}=0$ and incrementally updated using
\begin{equation}
    N_q^{(t)} = N_q^{(t-1)}  \gamma + \frac{n_q^{(t)}}{Bhw}  (1 - \gamma),
\end{equation}
where the number of encoded features assigned to the codebook entry $\bm{z}_q$ is represented by $n_q^{(t)}$. 
The hyperparameter $\gamma$ can take values between 0 and 1.
$B$ represents the batch size, $h$ the height and $w$ the width resulting in the total number of features $Bhw$.
To ensure that infrequently utilized codebook entries are updated more effectively, the decay value $\alpha_q^{(t)}$ is calculated for each codebook entry
\begin{equation}
    \alpha_q^{(t)} = \exp \left( -N_q^{(t)}  Q  \frac{10}{1 - \gamma} - \epsilon \right),
\end{equation}
where $Q$ is the total number of codebook entries and $\epsilon$ is a small constant that ensures the exponent remains negative.
This decay factor determines the degree of influence each codebook entry receives during the update process.
To update the codebook entries $\bm{z}_q$ toward the encoded feature vectors, so-called anchors $\hat{\bm{z}}_k$ are utilized. 
Anchors are determined by selecting the closest encoded feature vector to each codebook entry from the minibatch of the current training epoch.
The updated codebook entry $\bm{z}_q^{(t)}$ is then computed as a weighted combination of the previous codebook entry and the selected anchor
\begin{equation}
    \bm{z}_q^{(t)} = \bm{z}_q^{(t-1)}  (1 - \alpha_q^{(t)}) + \hat{\bm{z}}_q^{(t)}  \alpha_q^{(t)}.
\end{equation}
This update mechanism ensures that unused or rarely accessed codebook entries are gradually pulled closer to the embeddings of their nearest feature vectors. 
Improving the utilization of the latent space increases its information capacity. 
Zheng et al. \cite{Zheng2023} demonstrate that this enhanced utilization leads to improved reconstruction quality across multiple datasets.

\subsection{The Coupon Collectors Problem}
The \ac{CCP} is a probability problem that calculates the expected number of trials needed to collect all distinct items in a set, given that each item is randomly obtained with a certain probability. 
It represents a specific instance of the urn problem. 
The urn contains an unknown total number~$N$ of coupons, but the number of distinct coupon types~$j$ is known.
The objective is to determine the number of coupons that must be drawn independently with replacement to state with a given certainty $\tau$ that each coupon type is drawn at least once.
Each coupon type is drawn with a probability of \mbox{$p_j > 0$}, where \mbox{$1 \leq j \leq N$} and \mbox{$\sum_{j=1}^N p_j = 1$}. 
Let $X$ denote a random variable representing the number of draws required for all coupon types to be observed at least once.
Since an analytical solution for the expected value $\mathbb{E}_\textbf{X}$ is not available, \cite{Hauer2019} employs a Monte Carlo simulation to estimate the probability of observing all coupon types.
This involves repeatedly sampling~$S_i$~times from a random distribution until all coupon types have been observed at least once. 
Initially, this simulation is conducted $R$ times to estimate the mean $\overline{X}$ and standard deviation $\sigma$ for $X$.
The actual number of required simulations $sim$ can then be calculated
\begin{equation}
    sim \geq \frac{c_{1-\alpha/2}^2  \sigma^2}{e^2},
\end{equation}
where the hyperparameter $c_{1-\alpha/2}^2$ is the confidence level and $e$ denotes the standard error.
To assess the likelihood that all coupon types are drawn at least once within $Y$ samples, the probability $P(X \leq Y)$ is calculated. 
The first step to compute this is performing additional $I = sim - R$ simulations. 
For each of the $i=1,\dots,I$ simulations, $S_i$ is the number of samples needed to see all coupon types in the respective simulation.
Subsequently, the number of simulations in which the number of required samples $S_i$ was equal to $Y$ is determined.
This occurrence is defined as
\begin{equation}
    occ(i) = \sum_{i=1}^{I} \mathbf{1}_{S_i = Y},
\end{equation}
where $\mathbf{1}_{S_i = Y}$ is an indicator function that is 1 if the number of required samples $S_i$ is equal to $Y$, and 0 otherwise.
Dividing this sum by the total number of simulations $sim$, yields the desired probability \cite{Hauer2019}
\begin{equation}
    P(X \leq Y) = \frac{1}{sim}   occ(i).
\end{equation}
To determine the minimum number of samples $S_{\text{min}}$ required to meet a certainty threshold $\tau$, the occurrences $occ(i)$ are incrementally summed until their cumulative values is equal to or greater than $\tau$
\begin{equation}
S_{\text{min}} = \text{minimize } Y \text{ subject to } \frac{1}{sim} \sum_{i=1}^Y occ(i) \geq \tau.
\label{eq:minS}
\end{equation}
$\tau$ is the target value for $P(X \leq S_{\text{min}})$, ensuring that the cumulative probability of observing all coupon types reaches or exceeds the given certainty. 
Thereby, this approach provides insight into the  minimum number of samples $S_{\text{min}}$ needed to ensure all known coupon types are observed.

\section{Methodology}

This section introduces the novel pipeline for traffic scenario clustering and the completeness evaluation of the found categories, which is depicted in Fig. \ref{fig:CVQ-VAE_CCP_Concept_detail}.
After specifying the problem at hand, the pipeline is explained in detail. 
In the first step, the \ac{CVQ-VAE} is used to create varying codebooks with different numbers of traffic scenario categories.
In the second step, the impact of the number of traffic scenario categories on the required amount of data to assess a codebook as complete is evaluated.
For the completeness analysis, the~\ac{CCP} utilized. 

\definecolor{airforceblue}{rgb}{0.36, 0.54, 0.66}
\definecolor{amaranth}{rgb}{0.9, 0.17, 0.31}
\begin{figure*}[t]
    \centering
    \vspace{2em}
    \begin{tikzpicture}[node distance= 2cm, scale=0.95, , every node/.style={transform shape}]
    
        \tikzstyle{decision} = [diamond, draw, text centered, inner sep=5pt, line width = 1.3pt, font=\scriptsize]

        \node (K) [draw, rectangle, minimum width=12.5cm, minimum height=3.2cm, rounded corners = 4.4, line width = 1.0pt, fill=lightgray!15, font=\scriptsize] at (-3,-3.5) {}; 
        \node [below left = 0.2cm and -12.0cm of K, font=\large] {};
        \node (CVQ-VAE) [above left = -0.1cm and -2.1 cm of K, node distance = 2cm,   align=center, font = \large]{
        CVQ-VAE
        };

        \node (KW1) [below left = -2.25cm and -1.3cm of K, node distance= 5.0cm, draw, rectangle, minimum width=1.0cm, minimum height=1.0cm, align=center, line width = 0.5pt, font=\small] {};
        \node (KW2) [node distance= 0cm, fill=white, draw, rectangle, minimum width=1.0cm, minimum height=1.0cm, align=center, line width = 0.5pt, font=\scriptsize] at ($(KW1.south west) + (0.45,0.45)$) {};  
        \node (KW3) [node distance= 0cm, fill=white, draw, rectangle, minimum width=1.0cm, minimum height=1.0cm, align=center, line width = 0.5pt, font=\small] at ($(KW2.south west) + (0.45,0.45)$) {};

        \node (KW4) [below right = -2.25cm and -1.15cm of K, node distance= -5.1cm, draw, rectangle, minimum width=1.0cm, minimum height=1.0cm, align=center, line width = 0.5pt, font=\small] {};
        \node (KW5) [node distance= 0cm, fill=white, draw, rectangle, minimum width=1.0cm, minimum height=1.0cm, align=center, line width = 0.5pt, font=\scriptsize] at ($(KW4.south west) + (0.45,0.45)$) {};  
        \node (KW6) [node distance= 0cm, fill=white, draw, rectangle, minimum width=1.0cm, minimum height=1.0cm, align=center, line width = 0.5pt, font=\small] at ($(KW5.south west) + (0.45,0.45)$) {};
        
        \node (KE) [draw, trapezium, trapezium left angle=60, trapezium right angle=60, minimum width=2.2cm, minimum height=0.75cm, fill=cyan!10, line width = 1.0pt, rotate =-90, text centered, font=\large] at (-7.35,-3.5) {{\rotatebox{90}{E}}}; 
        \node [below right = -0.85 cm and -0.95cm of KE, font=\large] {$\bm{\xi}$};
        \node (KE2) [right of=KE,node distance= 1.0cm, draw, rectangle, minimum width=1.0cm, minimum height=1.0cm, fill=cyan!10, align=center, line width = 1.0pt, font=\Large]  {$\hat{\bm{z}}$};

        \node (argmin) [above right = -0.5cm and 0.15cm of KE2,  align=center, font = \small]{
        $\arg \min$\\$\left|\left| \hat{\bm{z}} - \bm{z}_q \right|\right|$
        };
        
        \node (KE3) [right of=KE,node distance= 7.7cm, draw, rectangle, minimum width=1.0cm, minimum height=1.0cm, fill=cyan!10, align=center, line width = 1.0pt, font=\Large]  {$\bm{z}_{q}$};
        \node (KD) [right of=KE3, node distance= 1.0cm, draw, trapezium, trapezium left angle=60, trapezium right angle=60, minimum width=2.2cm, minimum height=0.75cm, fill=cyan!10, line width = 1.0pt, rotate =90, text centered, font=\large] {{\rotatebox{-90}{D}}};
        \node [below right = 0.15 cm and 0.4cm of KD, font=\large] {$\bm{\hat{\xi}}$};

        \node (K_cl) [above right = 0.1cm and -3.22 cm of K, draw, rectangle, dashed, minimum width=3.2cm, minimum height=1.7cm, rounded corners = 4.4, line width = 1.0pt, fill=lightgray!15, font=\scriptsize] {}; 
        \node (Classifier) [ draw, trapezium, trapezium left angle=60, trapezium right angle=60, minimum width=1.2cm, minimum height=0.5cm, fill=cyan!10, line width = 1.0pt, rotate =90, align=center] at (1.3,-1) {{\rotatebox{-90}{$f_{\text{cl}}$}}};
        \draw[->, line width = 1.0pt] (KE3) |- (Classifier.north);
        \node (softmax) [right of=Classifier,node distance = 0.7cm, draw, rectangle, minimum width=0.1cm, minimum height=1.4cm, fill=white,  line width = 0.7pt, font=\small] {{\rotatebox{90}{softmax}}};

        \draw[->, line width =1.0pt] (Classifier)--++ (1.25,0);
        \node (softmax) [right of=Classifier,node distance = 0.7cm, draw, rectangle, minimum width=0.1cm, minimum height=1.4cm, fill=white,  line width = 0.7pt, font=\small] {{\rotatebox{90}{softmax}}};
        \node (pcl) [right of = softmax, node distance = 0.9 cm, font=\large] {$\bm{p}_{\text{cl}}$};
        
        \node (man_clas) [above left = -1.0cm and 0 cm of K_cl, node distance = 2cm,   align=right, font = \large]{
        Future Behavior\\Prediction
        };

        \node (QB) [draw, rectangle, minimum width=3cm, fill=white, minimum height=3.0cm, align=center, rounded corners = 3.4, line width = 1.0pt, font=\scriptsize] at (-2.5,-3.5) {};
        \node [above = -0.55cm of QB, font=\footnotesize] {Codebook $\mathit{Q}$ };

        \node (QB2) [above right = -1cm and -2.9cm of QB, draw, rectangle, minimum width=0.41cm, minimum height=0.41cm, fill=white, align=center, line width = 0.7pt, font=\footnotesize] {1};
        \node (QB21) [right = 0.0cm of QB2,draw, rectangle, minimum width=2.3cm, minimum height=0.41cm, fill=lime!15, align=center, line width = 0.7pt, font=\footnotesize]  {$\bm{z}_{1} \quad \rightarrow \quad  p_1$};
        \node (QB3) [above right = -0.87cm and -0.43cm of QB2, draw, rectangle, minimum width=0.41cm, minimum height=0.41cm, fill=white, align=center, line width = 0.7pt, font=\footnotesize]{2};
        \node (QB31) [right = 0.0cm of QB3,draw, rectangle, minimum width=2.3cm, minimum height=0.41cm, fill=lime!40, align=center, line width = 0.7pt, font=\footnotesize]  {$\bm{z}_{2} \quad \rightarrow \quad p_2$};
        \node (QB4) [above right = -0.87cm and -0.43cm of QB3, draw, rectangle, minimum width=0.41cm, minimum height=0.41cm, fill=white, align=center, line width = 0.7pt, font=\footnotesize] {3};
        \node (QB41) [right = 0.0cm of QB4,draw, rectangle, minimum width=2.3cm, minimum height=0.41cm, fill=lime, align=center, line width = 0.7pt, font=\footnotesize]  {$\bm{z}_{3} \quad \rightarrow  \quad p_3$};
        \node (QB5) [above right = -1.5cm and -0.43cm of QB4, draw, rectangle, minimum width=0.41cm, minimum height=0.41cm, fill=white, align=center, line width = 0.7pt, font=\footnotesize] {$Q$};
        \node (QB51) [right = 0.0cm of QB5,draw, rectangle, minimum width=2.25cm, minimum height=0.41cm, fill=lime!55, align=center, line width = 0.7pt, font=\footnotesize]  {$\bm{z}_{Q} \quad \rightarrow \quad  p_Q$};
        % ...
        \node [below right = -0.2cm and 1.0cm of QB4, font=\large] {\vdots };

        \draw[->, line width=1.0pt] (KE2) -- (QB);
        \draw[->, line width=1.0pt] (QB) -- (KE3);


        \node (K_cl) [right =  0.1 cm of K, draw, rectangle, minimum width=5.7cm, minimum height=3.2cm, rounded corners = 4.4, line width = 1.0pt, fill=lightgray!15, font=\scriptsize] {}; 

        \node (Q_new) [above right = -1.6cm and 0.25cm of K, draw, rectangle, fill= airforceblue!70, minimum height=0.5cm, minimum width = 1.8 cm, line width = 0.7pt, rounded corners = 2.4, font=\large] {$ p_{\text{new}}$};

        \node (tau) [ below = 0.3cm of Q_new, draw, rectangle, fill=amaranth!40, minimum height=0.5cm, minimum width = 1.8 cm, line width = 0.7pt, rounded corners = 2.4, font=\large] {$\tau$};

        \node (CC) [right =  2.5cm of K, draw, rectangle, fill = white, minimum height=2cm, minimum width = 2.1 cm, line width = 0.7pt, align =center, font=\large] {};

        \node (S_min) [right = 0.3cm of CC,  font=\large] {$S_{\text{min}}$};

        \node (Completeness Check) [above left = -0.1cm and -3.9 cm of K_cl, node distance = 2cm,   align=center, font = \large]{
        Completeness Check
        };

       \foreach \i [count=\n] in {1.1, 1.0, 0.9, 0.8, 0.7, 0.6, 0.5} {
        \fill[lime!80] (5.85+\n*0.2,-4.2) rectangle (5.85+\n*0.2+0.15, -4.2+\i);
        }
        \fill[airforceblue!85] (7.45,-4.2) rectangle (7.6,-3.8);
        \draw[-, line width = 1pt,dashed, color = amaranth!80] (6.0, -3.3) -- (7.7, -3.3);
        \draw[->, line width = 1.5pt] (6.0,-4.23) --(6.0,-2.9);
        \draw[->, line width = 1.5pt] (6.0, -4.2) --(7.8,-4.2);

        \path [->, line width =1.0pt]
        (tau) edge (tau-|CC.west)
        (Q_new) edge (Q_new-|CC.west)
        (CC) edge (CC -| S_min.west)     
        ;

        \draw[->, line width=1.0pt ] (QB.east) --++ (0.4,0) --++ (0,-1.3) -|(CC.south) ;

    \end{tikzpicture}
    \vspace{-5pt}
    \caption{\small The proposed pipeline in detail consists of: the CVQ-VAE, a future behavior predictor and the completeness check. The CVQ-VAE discretizes the driving scenario $\bm{\xi}$. Its performance is improved by the future behavior predictor. The codebook~$Q$ created in the process contains the various driving scenario categories~$q$ and their probabilities of occurrence~$p_q$. The completeness check uses these category probabilities~$p_q$ and the hyperparameters~$p_{\text{new}}$ and~$\tau$ to compute the required amount of data~$S_{\text{min}}$ needed to achieve scenario category completeness with confidence $\tau$.}
    \vspace{-10pt}
    \label{fig:CVQ-VAE_CCP_Concept_detail}
\end{figure*}
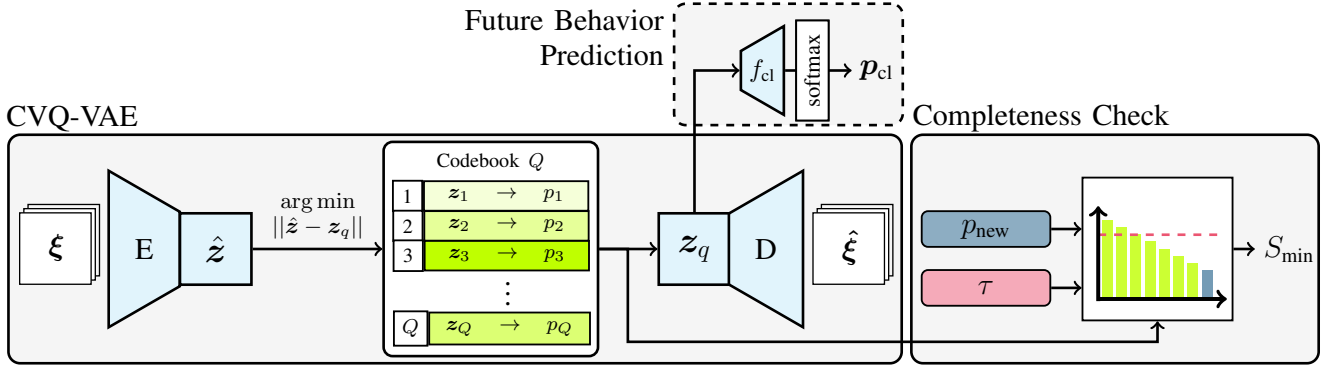 

\subsection{Problem Description}
Based on a dataset \mbox{$\mathcal{D} = \{(\boldsymbol{\bm{\xi}}^{(m)}, s^{(m)})\}_{m=1}^M$} consisting of $M$ samples, this work aims to identify a total of $Q$ representative traffic scenario categories. 
Each data sample consists of the traffic scenario $\boldsymbol{\bm{\xi}}^{(m)} \in \mathbb{R}^{N \times F \times T_{\text{obs}}}$ and the future behavior class $s^{(m)} \in \mathbb{R}^S$.
Each traffic scenario $\boldsymbol{\bm{\xi}}^{(m)}$ contains $F$~vehicle features for up to a total number of $N$ vehicles.
For each $j-$th participating vehicle, the features $F$ comprise the longitudinal and lateral positions $(x_j,y_j)$ as well as velocities $(v_{j,\text{x}},v_{j,\text{y}})$ for $T_{\text{obs}}$ time steps.
$s^{(m)}$ contains information on the future movement of the target vehicle in the form of a one-hot encoding.
The objective is to utilize the data to train a model $g$, performing 
\begin{equation}
    g : \left( \xi^{(m)}, s^{(m)} \right) \mapsto q,
\end{equation}
which assigns each traffic scenario to a traffic scenario category $q= 1, \dots, Q$.
The information given by the future behavior class is utilized to enrich the latent space and to evaluate the results.
After clustering the traffic scenarios, the resulting traffic scenario catalogs are evaluated for their  scenario category completeness.

\subsection{Model Architecture for Traffic Scenario Clustering} 
For the clustering step the \mbox{\ac{CVQ-VAE}} model is employed. 
The \mbox{\ac{CVQ-VAE}} is a specialized form of autoencoder, consisting of an encoder~$\text{E}$ and a decoder~$\text{D}$, where the latent space is discretized using \ac{VQ}.
The input traffic scenario $\boldsymbol{\bm{\xi}}^{(m)}$ is mapped by an encoder $\text{E}$ to a latent representation $\hat{\bm{z}}^{(m)} \in \mathbb{R}^{R_q}$, $\text{E}:(\boldsymbol{\bm{\xi}}^{(m)}) \mapsto \hat{\bm{z}}^{(m)}$. 
The key aspect of this approach lies in the treatment of the latent representations. 
While the classical \ac{VAE} considers the prior and posterior distributions as continuous, the \ac{CVQ-VAE} operates with discrete, categorical distributions.
This approach is particularly suitable for real-world categorical data~\cite{Oord2017}. 
Values drawn from the discrete categorical distribution are used to select an embedding $q$, referred to as codebook entry, from a predefined embedding table called codebook with $q \in \{ 1,...,Q \}$ and $Q$ being the total number of codebook entries.
To select a codebook entry, $\hat{\bm{z}}^{(m)}$ is discretized as $\bm{z}_q^{(m)} = f_q\left(\hat{\bm{z}}^{(m)}\right)$. 
This is achieved by calculating the Euclidean distance between $\hat{\bm{z}}^{(m)}$ and each codebook vector in $\mathcal{Z} = { \bm{z}_1, \dots, \bm{z}_Q}$, which hold the latent features to their belonging codebook entry $q$:
\begin{equation} 
\boldsymbol{\bm{z}}{_q}^{(m)} = f_q\left(\hat{\bm{z}}^{(m)}\right) = \arg \min_{\bm{z_q} \in \mathcal{Z}} \left|\left| \hat{\bm{z}}^{(m)} - \bm{z}_q \right|\right|_2^2, 
\end{equation}
where $\bm{z}_q \in \mathbb{R}^{R_q}$.
The codebook vector with the smallest Euclidean distance, i.e., the traffic scenario category which represents the traffic scenario $(m)$ best, is selected. 
Then $\bm{z}_q$ serves as input to the decoder which tries to reconstruct the original data, $\text{D}:(\bm{z}_q) \mapsto \hat{\boldsymbol{\xi}}^{(m)}$ with $\hat{\boldsymbol{\xi}}^{(m)} \in \mathbb{R}^{N \times F \times T_{\text{obs}}}$~\cite{Oord2017}.
To train the model, the loss function 
\begin{equation}
    \begin{aligned}
        \mathcal{L}_{\text{cvq}} = \| \bm{\xi}^{(m)} - \hat{\bm{\xi}}^{(m)} \|^2_2 
        &+ \| \text{sg}[\text{E}(\bm{\xi}^{(m)})] - \bm{z}_q^{(m)} \|^2_2 \\
        &+ \beta \| \text{sg}[\bm{z}_q^{(m)}] - \text{E}(\bm{\xi}^{(m)}) \|^2_2,
    \end{aligned}
\end{equation}
is employed. 
The first term denotes reconstruction loss.
By minimizing it, the model learns to reconstruct the data correctly, thereby optimizing both the encoder and the decoder.
The second term, the vector quantization loss binds the codebook entries $\bm{z}_q$ to the latent vectors \mbox{$\text{E}(\bm{\xi}^{(m)}) = \hat{\bm{z}}^{(m)}$} generated by the encoder.
$sg[\cdot]$ denotes the stop-gradient operator, which treats its argument as a constant during backpropagation. 
The last term, denoted as commitment loss, is weighted with $\beta$ and ensures that the encoder output does not deviate strongly from the codebook entries.

Following the quantization step, an additional implemented classifier assigns the selected codebook entries to one of $S$ predetermined future behavior classes.
A simple linear classifier with a cross-entropy loss is used to penalize misclassifications
\begin{equation}
\mathcal{L}_{\text{cl}} = - \sum_{i=1}^{S} s_i^{(m)}  \log(p_i^{(m)}),
\end{equation}
\begin{equation}
p_{\text{cl}}^{(m)} = \text{softmax}(f_{\text{cl}}(\hat{\bm{z}}^{(m)})),
\end{equation}
where $s^{(m)}$ is one of the ground truth future behavior classes.
The predicted label is described by \mbox{$p_{\text{cl}}^{(m)} \in \mathbb{R}^S$~\cite{Neumeier2024}}. 
This addition ensures that scenarios with different target vehicle behaviors are distinctly grouped into different traffic scenario categories.
This classification loss is weighted with $\lambda$ and added to $\mathcal{L}_{\text{cvq}}$ resulting in the total loss
\begin{equation}
    \mathcal{L}_{\text{total}} = \mathcal{L}_{\text{cvq}} + \lambda \mathcal{L}_{\text{cl}}.
\end{equation}
By adjusting the weight~$\lambda$, the influence of the classifier on clustering behavior can be analyzed. 
Training models with varying predefined numbers of codebook entries $Q$ allows the creation of codebooks of varying size.
Building on these codebooks the relative frequency~$p_q$ of each codebook entry~$\bm{z}_q$ can be determined by counting the number of scenarios assigned to that entry:
\begin{equation} 
    p_q = \frac{n_q}{\sum_{q=1}^Q n_q}, 
    \label{eq:p_q}
\end{equation}
where $n_q$ denotes the total number of scenarios associated with codebook entry $\bm{z}_q$ and $Q$ is the total number of codebook entries.

\subsection{Completeness of Traffic Scenario Categories} 
The Coupon Collectors Problem (\ac{CCP}) is applied to analyze the requirements a dataset must fulfill to be considered complete, following the approach of Hauer et al.~\cite{Hauer2019}.
In this context, the coupon types represent the traffic scenario categories.
There is a total of $Q$ distinct coupon types.
The number of samples corresponds to the individual traffic scenarios $\boldsymbol{\bm{\xi}}^{(m)}$, that must be included in the dataset $\mathcal{D}$ for it to be described as complete.
Up to this point, this method determines how many traffic scenarios are required to ensure, with a confidence level of $\tau$, that all categories of traffic scenarios already known are included within $\mathcal{D}$.
However, to estimate the completeness, it is essential to determine the number of required scenarios needed to detect a previously unknown scenario category.
This is analog to a traffic scenario category that does not occur in the collected dataset.
Therefore, a new category with probability $p_{\text{new}}$ is introduced. 
The already known probabilities are subsequently linearly scaled to $p_q'$ such that ${p_{\text{new}} + \sum_{q=1}^Q p'_q = 1}$.
In the subsequent experiment, sampling with replacement is performed from the available categories until every traffic scenario category, including the newly introduced one, has been drawn at least once. 
This process simulates the case of observing all categories in a dataset. 
The experiment is repeated $sim$ times to account for variability and to ensure statistical robustness. 
The number of samples $S_{\text{min}}$ required to guarantee, with a confidence level of $\tau$, that all categories have been observed can be calculated using \mbox{Equation (\ref{eq:minS})}. 
By reversing this reasoning, it can be inferred that if a dataset contains $S_{\text{min}}$ traffic scenarios, it includes, with a confidence level of $\tau$, all traffic scenario categories that occur with a frequency greater than $p_{\text{new}}$. 
This provides an assessment of the dataset completeness in terms of categorical coverage, ensuring both theoretical rigor and practical applicability.

\renewcommand{\footnoterule}{%
  \kern -3pt
  \hrule width 4cm height 0.5pt
  \kern 2pt
}

\section{Experiments and Results}

This section outlines the dataset used and the applied preprocessing steps.
Subsequently, the implementation details are described.
In the next step, the clustering results are presented.
Based on these results, the completeness considerations are analyzed.

\subsection{Dataset}

The dataset used in this study is the publicly available highD dataset \cite{Krajewski2018}, which consists of recordings of German highway segments captured from a drone at \SI{25}{\Hz}. 
The preprocessing steps follow those outlined in \cite{Neumeier2024, Neumeier2021}.
Each extracted scenario $\bm{\xi}^{(m)}$ contains up to $N = 9$ vehicles. 
These are the target vehicle and the up to eight surrounding vehicles.
For each vehicle $F = 4$ features are stored for $T_{\text{obs}} = 75$ time steps \mbox{(\SI{3}{s})}. 
The vehicle features include the longitudinal ($\textbf{x}$) and lateral ($\textbf{y}$) positions as well as the corresponding velocities $\bm{v}_\text{x}$ and $\bm{v}_\text{y}$. 
Each scenario was categorized into one of \mbox{$S =3$} classes: \ac{kl}, \ac{lcl}, or \ac{lcr}, depending on the subsequent driving behavior of the target vehicles. 
This class information offers additional context regarding the observed traffic scenario. 
The dataset was balanced to ensure that an approximately equal number of traffic scenarios is available for each of the three pseudo-classes. 
The training dataset $\mathcal{D}_{\text{train}}$ contains a total of 9,841 samples, while the test dataset $\mathcal{D}_{\text{test}}$ includes 4,217 samples.

\subsection{Implementation Details}
The \ac{CVQ-VAE} was trained with a batch size of \mbox{$B = 64$} until convergence of the reconstruction loss. 
This corresponded to a total of \mbox{$E = 1,500$} epochs.
A learning rate of $\alpha = 0.001$ was used and the weight $\lambda$ was set to $0.2$. 
The commitment loss was assigned a constant weight of \mbox{$\beta = 0.25$}.
The latent space feature dimension $R_\text{q}$ is set to 64.
Therefore, each codebook entry has a dimension of \mbox{$\bm{z}_q \in \mathbb{R}^{64}$}.
Three different models were trained, each with a different number of codebook entries: $Q_1 = 64$, $Q_2 = 128$ and $Q_3 = 256$.
In line with the experiments in~\cite{Hauer2019}, the confidence level is chosen to be \mbox{$ c_{1-0.05/2}^2 = 1.96$} and the standard error to be \mbox{$e= 0.01$}.
The experiments were repeated for each model with the three probabilities \mbox{$p_{\text{new},1} = 0.001$}, \mbox{$p_{\text{new},2} = 0.0001$}  and \mbox{$p_{\text{new},3} = 0.00001$}, with the \mbox{threshold $\tau_1 = 0.95$.} 
For details of architecture implementation and code see: \url{https://github.com/mb-team-thi/completeness-pipeline}.

\subsection{Traffic Scenario Clustering}

\begin{table}[t]
    \centering
    \vspace{1em}
\begin{tabularx}{\linewidth}{c|c|c|c|c|c}

    \makecell{Model} & 
    \makecell{Codebook\\Usage $\uparrow$} & 
    \makecell{$\mathcal{L}_{\text{R,train}}$\\$\downarrow$} & 
    \makecell{$\mathcal{L}_{\text{R,test}}$\\$\downarrow$} &
    \makecell{$H_{\text{avg,train}}$\\$\downarrow$} &
    \makecell{$H_{\text{avg,test}}$\\$\downarrow$} \\ \toprule \toprule

    VQ-VAE \cite{Neumeier2024} & 49/60 & - & - & 0.01 & 0.39 \\ 
    \midrule
    CVQ-VAE & 64/64 & 0.41 & 4.40 & 0.042 & 0.56 \\ 
    CVQ-VAE & \textbf{128/128} & 0.32 & 3.50 & 0.021 & 0.53 \\ 
    CVQ-VAE & 253/256 & \textbf{0.23} & \textbf{2.70} & \textbf{0.004} & \textbf{0.35} \\ 

\end{tabularx}

    \caption{Ablation study showing the influence of the number of codebook entries on $\mathcal{L}_\text{R}$  and Shanon entropy $H_{\text{avg}}$ on the train and test set.}
    \label{tab:Ablation_Study}
    \vspace{-1.5em}
\end{table}

\begin{figure} [t]
    \centering
    \begin{subfigure}[b]{\linewidth}
	\centering
	\vspace{1pt}
	\begin{tikzpicture}
	\begin{axis}[
	ybar stacked,
	ymajorgrids,
	grid style={dashed,gray!30},
	bar width = 3pt,
	height=3.0cm,
	xtick align=center,
	width=0.95*\linewidth,
	xlabel={\small $q$},
	ylabel={\small $\log_2(\#)$},
	x label style={at={(axis description cs:0.5,-0.3)}, anchor=north},
	y label style={at={(axis description cs:-0.05,.5)}, anchor=south},
	xtick={0,10,20,30,40,50,60},
	xticklabels={0,10,20,30,40,50,60},
	table/col sep=comma,
	ymin=0,
	xmin=-0.5,
	xmax=60,
	legend entries={ \small lcl, \small kl,\small lcr},
	legend style={at={(1,0.5)}, xshift = 0.55cm, 
		anchor=center, nodes=right,  minimum size=0.2cm, inner sep=1pt},
	legend columns=1,
	]
	\addplot+ table[x =idx, y=lcl] {data/e1062_train_loghist.csv};
	\addplot+ table[x =idx, y=kl] {data/e1062_train_loghist.csv};
	\addplot+ table[x =idx, y=lcr] {data/e1062_train_loghist.csv};
	\end{axis}
	\end{tikzpicture}
    \vspace{-7pt}
    \caption{$Q$ = 60 (VQ-VAE) \cite{Neumeier2024}}
    \vspace{2pt}
\end{subfigure}
\begin{subfigure}[b]{\linewidth}
	\centering
	\vspace{1pt}
	\begin{tikzpicture}
	\begin{axis}[
	ybar stacked,
	ymajorgrids,
	grid style={dashed,gray!30},
	bar width = 3pt,
	height=3.0cm,
	xtick align=center,
	width=0.95*\linewidth,
	xlabel={\small $q$},
	ylabel={\small $\log_2(\#)$},
	x label style={at={(axis description cs:0.5,-0.3)}, anchor=north},
	y label style={at={(axis description cs:-0.05,.5)}, anchor=south},
	xtick={0,10,20,30,40,50,60},
	xticklabels={0,10,20,30,40,50,60},
	table/col sep=comma,
	ymin=0,
	xmin=-0.5,
	xmax=64,
	legend entries={ \small lcl, \small kl,\small lcr},
	legend style={at={(1,0.5)}, xshift = 0.55cm, 
		anchor=center, nodes=right,  minimum size=0.2cm, inner sep=1pt},
	legend columns=1,
	]
	\addplot+ table[x =idx, y=lcl] {data/output_data_64.csv};
	\addplot+ table[x =idx, y=kl] {data/output_data_64.csv};
	\addplot+ table[x =idx, y=lcr] {data/output_data_64.csv};
	\end{axis}
	\end{tikzpicture}
    \vspace{-7pt}
    \caption{$Q$ = 64 (CVQ-VAE)}
    \vspace{2pt}
\end{subfigure}
\caption{Stacked histogram showing the codebook usage in (a) the work \cite{Neumeier2024} compared to (b) this study for the codebook size $Q_1 =64$. Colors indicate the ground truth future behavior class. The histogram has a logarithmic scale.}
    \label{fig:Codebook_Usage}
    \vspace{-15pt}
\end{figure}
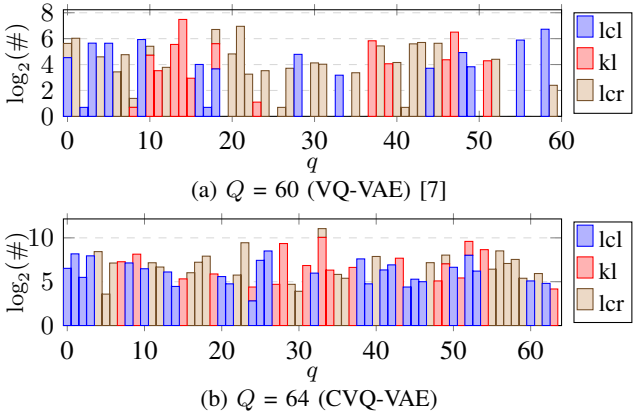

After completing the training of the \ac{CVQ-VAE} the performance is evaluated with different metrics. 
As Table \ref{tab:Ablation_Study} shows these metrics are the codebook usage, the cluster purity expressed by $H_{\text{avg}}$ and the reconstruction quality.
First the codebook usage is considered.
Therefore, the assignment of each traffic scenario $\bm{\xi}^{(m)}$ in $\mathcal{D}_{\text{train}}$ to its respective codebook entry $q$ was recorded. 
While the \ac{VQ-VAE} architecture in \cite{Neumeier2024} was only capable of utilizing 49 out of 60 codebook entries, this study achieved a remarkable improvement, with all 64 codebook entries being effectively utilized for \mbox{$Q_1 = 64$}.
Similarly, the \ac{CVQ-VAE} successfully made use of all 128 codebook entries for \mbox{$Q_2 = 128$}.
However, for \mbox{$Q_3 = 256$}  only 253 entries are utilized, suggesting a diminished usage as the number of codebook entries increases.
Additional experiments with higher numbers of codebook entries have shown that the proportion of unused codebook entries increases as the total number of entries grows.
Thus, while the architecture in this study is designed to enforce the use of all codebook entries, this enforcement becomes less practical when the number of entries is too large.
The exact reasons for this behavior and potential strategies to achieve full utilization of codebook entries, regardless of their total number, require further investigation.
Nevertheless, the \ac{CVQ-VAE} successfully improves codebook usage. 
Fig.~\ref{fig:Codebook_Usage} illustrates the codebook usage for  \mbox{$Q_1 = 64$} in comparison to the codebook usage in~\cite{Neumeier2024}.
The scenarios ground truth future behavior class \mbox{(\ac{kl}, \ac{lcl}, \ac{lcr})}, is indicated by the color of the bar.
These are used to evaluate the quality of the formed traffic scenario categories.
The core idea is that scenarios belonging to the same category share a similar traffic context, resulting in comparable scenario outcomes.
If multiple future behavior classes are present within a single scenario category, it may suggest that the category contains strongly differing scenarios.
To determine the degree of future behavior class variation within the codebook entries, the average Shannon entropy \mbox{$H_{\text{avg}} = \mathbb{E}_{q=1...Q}[H_q]$} over the entire codebook is calculated~\cite{Neumeier2024}. 
The entropy $H_q$ for each codebook entry $q$ is
\begin{equation}
    H_q = - \sum_{i=1}^S p_{\text{cl},i}  \log_2  p_{\text{cl},i},
\end{equation}
where $p_i$ is the predicted probability for each future behavior class $S_i$ for the latent scenario representation~$\hat{z}^{(m)}$.
The Shannon entropy can indicate complete purity \mbox{($H_q = 0$)} up to complete impurity \mbox{($H_q = \log_2(3)= 1.585$)}, since \mbox{$S = 3$}.
The calculations were conducted for all three models variations for both $\mathcal{D}_{\text{train}}$ and $\mathcal{D}_{\text{test}}$.
Table \ref{tab:Ablation_Study} shows that with an increasing number of codebook entries, the average purity of the individual codebook entries improves strongly.
\begingroup
\setlength{\tabcolsep}{3pt}  
\renewcommand{\arraystretch}{0.95}  

\begin{figure}[t]
    \centering
    \begin{subfigure}[b]{0.3\linewidth}
        
        \footnotesize
        \centering
        
        \vspace{2.5em} 
        \begin{tabular}{@{}c@{}c@{}}
        \raisebox{-2em}{\rotatebox{90}{\small true class}}
        \hspace{0.2em}
        &
        \begin{tabular}{c|ccc}
            & lcl & kl & lcr \\
            \hline
            lcl & \cellcolor{lime!40}0.87 & 0.12 & 0.01 \\
            kl  & 0.11 & \cellcolor{lime!40}0.85 & 0.04 \\
            lcr & 0.01 & 0.12 & \cellcolor{lime!40}0.87 \\
        \end{tabular}
        
    \end{tabular}
    \caption{$Q_1 =64$}
    \end{subfigure}
    \hfill
    \hspace{0.6em}
    \begin{subfigure}[b]{0.3\linewidth}
        \centering
        \footnotesize
        
        \vspace{0.5em}
        {\small predicted class}
        \vspace{1em}

        \begin{tabular}{c|ccc}
            & lcl & kl & lcr \\
            \hline
            lcl & \cellcolor{lime!40}0.85 & 0.14 & 0.01 \\
            kl  & 0.09 & \cellcolor{lime!40}0.86 & 0.05 \\
            lcr & 0.01 & 0.11 & \cellcolor{lime!40}0.88 \\
        \end{tabular}
        \caption{$Q_2 = 128$}
    \end{subfigure}
    \hfill
    \begin{subfigure}[b]{0.3\linewidth}
        \centering
        \footnotesize
        \vspace{2.5em}
        \begin{tabular}{c|ccc}
            & lcl & kl & lcr \\
            \hline
            lcl & \cellcolor{lime!40}0.86 & 0.12 & 0.02 \\
            kl  & 0.09 & \cellcolor{lime!40}0.85 & 0.06 \\
            lcr & 0.01 & 0.09 & \cellcolor{lime!40}0.90 \\
        \end{tabular}
        \caption{$Q_3 = 253$}
    \end{subfigure}
    
    \caption{Confusion matrices for model $Q_1$, $Q_2$ and $Q_3$ showing the class prediction accuracy. }
    \vspace{2 pt}
    \label{fig:confusion_matrices}
\end{figure}
\endgroup

Similarly to \cite{Neumeier2024}, the models exhibit a discrepancy between $H_{\text{avg,train}}$ and $H_{\text{avg,test}}$.
This indicates that also the \mbox{\ac{CVQ-VAE}} still has potential for future improvements.
In direct comparison to \cite{Neumeier2024}, the \ac{CVQ-VAE} demonstrates superior performance only for a higher number of codebook entries, as shown in Table~\ref{tab:Ablation_Study}.
This can be partially attributed to the lower weighting $\lambda$ of the classifier module compared to \cite{Neumeier2024}, which influences the purity.
Table \ref{tab:Ablation_Study} also presents $\mathcal{L}_{\text{R,train}}$ and $\mathcal{L}_{\text{R,test}}$, the reconstruction losses, which notably decrease as the number of codebook entries increases.
This indicates that a lower reconstruction loss, achieved by increasing the number of codebook entries, reflects a closer resemblance between the input scenarios and their corresponding representatives. 
As a result, this improvement implies the formation of more coherent and well-defined clusters.
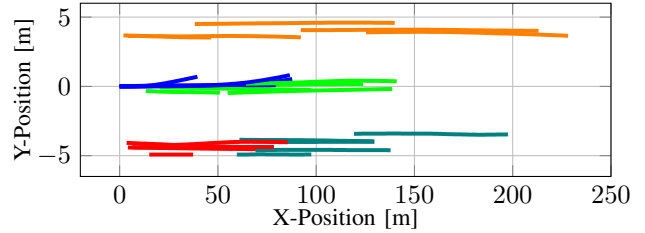
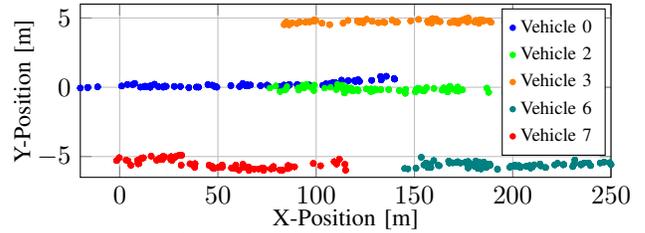
\begin{figure} [t]
    \centering
    \begin{subfigure}[b]{\linewidth}

        \begin{tikzpicture}
        \begin{axis}[
        width=\linewidth, height=0.45\linewidth,
        ymin =-6.5, ymax= 6,
        xmin= -20, xmax = 250,
        xlabel={\small X-Position [m]},
        ylabel={\small Y-Position [m]},
        x label style={at={(axis description cs:0.5,-0.13)}, anchor=north},
	y label style={at={(axis description cs:-0.07,.5)}, anchor=south},
        grid=major,
        legend pos=north east,
        ]

        \addplot[scatter,line width = 1.5pt, no marks, draw = blue]
        table [x=x_vehicle_0, y=y_vehicle_0, col sep = comma] {data/scenario_lcl2648_trajectories_transformed.csv};
        
        \addplot[scatter,line width = 1.5pt, no marks, draw = green]
        table [x=x_vehicle_2, y=y_vehicle_2, col sep = comma] {data/scenario_lcl2648_trajectories_transformed.csv};

        \addplot[scatter,line width = 1.5pt, no marks, draw = orange]
        table [x=x_vehicle_3, y=y_vehicle_3, col sep = comma] {data/scenario_lcl2648_trajectories_transformed.csv};

        \addplot[scatter,line width = 1.5pt, no marks, draw = teal]
        table [x=x_vehicle_6, y=y_vehicle_6, col sep = comma] {data/scenario_lcl2648_trajectories_transformed.csv};

        \addplot[scatter,line width = 1.5pt, no marks, draw = red]
        table [x=x_vehicle_7, y=y_vehicle_7, col sep = comma] {data/scenario_lcl2648_trajectories_transformed.csv};

        \addplot[scatter,line width = 1.5pt, no marks, draw = blue]
        table [x=x_vehicle_0, y=y_vehicle_0, col sep = comma] {data/scenario_lcl2081_trajectories_transformed.csv};
        
        \addplot[scatter,line width = 1.5pt, no marks, draw = green]
        table [x=x_vehicle_2, y=y_vehicle_2, col sep = comma] {data/scenario_lcl2081_trajectories_transformed.csv};

        \addplot[scatter,line width = 1.5pt, no marks, draw = orange]
        table [x=x_vehicle_3, y=y_vehicle_3, col sep = comma] {data/scenario_lcl2081_trajectories_transformed.csv};

        \addplot[scatter,line width = 1.5pt, no marks, draw = teal]
        table [x=x_vehicle_6, y=y_vehicle_6, col sep = comma] {data/scenario_lcl2081_trajectories_transformed.csv};

        \addplot[scatter,line width = 1.5pt, no marks, draw = red]
        table [x=x_vehicle_7, y=y_vehicle_7, col sep = comma] {data/scenario_lcl2081_trajectories_transformed.csv};

        \addplot[scatter,line width = 1.5pt, no marks, draw = blue]
        table [x=x_vehicle_0, y=y_vehicle_0, col sep = comma] {data/scenario_lcl1321_trajectories_transformed.csv};
        
        \addplot[scatter,line width = 1.5pt, no marks, draw = green]
        table [x=x_vehicle_2, y=y_vehicle_2, col sep = comma] {data/scenario_lcl1321_trajectories_transformed.csv};

        \addplot[scatter,line width = 1.5pt, no marks, draw = orange]
        table [x=x_vehicle_3, y=y_vehicle_3, col sep = comma] {data/scenario_lcl1321_trajectories_transformed.csv};

        \addplot[scatter,line width = 1.5pt, no marks, draw = teal]
        table [x=x_vehicle_6, y=y_vehicle_6, col sep = comma] {data/scenario_lcl1321_trajectories_transformed.csv};

        \addplot[scatter,line width = 1.5pt, no marks, draw = red]
        table [x=x_vehicle_7, y=y_vehicle_7, col sep = comma] {data/scenario_lcl1321_trajectories_transformed.csv};

        \addplot[scatter,line width = 1.5pt, no marks, draw = blue]
        table [x=x_vehicle_0, y=y_vehicle_0, col sep = comma] {data/scenario_lcl3133_trajectories_transformed.csv};
        
        \addplot[scatter,line width = 1.5pt, no marks, draw = green]
        table [x=x_vehicle_2, y=y_vehicle_2, col sep = comma] {data/scenario_lcl3133_trajectories_transformed.csv};

        \addplot[scatter,line width = 1.5pt, no marks, draw = orange]
        table [x=x_vehicle_3, y=y_vehicle_3, col sep = comma] {data/scenario_lcl3133_trajectories_transformed.csv};

        \addplot[scatter,line width = 1.5pt, no marks, draw = teal]
        table [x=x_vehicle_6, y=y_vehicle_6, col sep = comma] {data/scenario_lcl3133_trajectories_transformed.csv};

        \addplot[scatter,line width = 1.5pt, no marks, draw = red]
        table [x=x_vehicle_7, y=y_vehicle_7, col sep = comma] {data/scenario_lcl3133_trajectories_transformed.csv};

        \addplot[scatter,line width = 1.5pt, no marks, draw = blue]
        table [x=x_vehicle_0, y=y_vehicle_0, col sep = comma] {data/scenario_lcl495_trajectories_transformed.csv};
        
        \addplot[scatter,line width = 1.5pt, no marks, draw = green]
        table [x=x_vehicle_2, y=y_vehicle_2, col sep = comma] {data/scenario_lcl495_trajectories_transformed.csv};

        \addplot[scatter,line width = 1.5pt, no marks, draw = orange]
        table [x=x_vehicle_3, y=y_vehicle_3, col sep = comma] {data/scenario_lcl495_trajectories_transformed.csv};

        \addplot[scatter,line width = 1.5pt, no marks, draw = teal]
        table [x=x_vehicle_6, y=y_vehicle_6, col sep = comma] {data/scenario_lcl495_trajectories_transformed.csv};

        \addplot[scatter,line width = 1.5pt, no marks, draw = red]
        table [x=x_vehicle_7, y=y_vehicle_7, col sep = comma] {data/scenario_lcl495_trajectories_transformed.csv};

    \end{axis}
    \end{tikzpicture}
    \vspace{-8pt}
    \caption{Selected vehicle trajectories of scenarios assigned to the codebook entry \mbox{$q =167$}.}
    \vspace{11pt}
    \label{fig:sub1}
    
\end{subfigure}

\begin{subfigure}[b]{\linewidth}
    
    \begin{tikzpicture}
    \begin{axis}[
        width=\linewidth, height=0.45\linewidth,
        ymin =-6.5, ymax= 6,
        xmin= -20, xmax = 250,
        xlabel={\small X-Position [m]},
        ylabel={\small Y-Position [m]},
        x label style={at={(axis description cs:0.5,-0.13)}, anchor=north},
	y label style={at={(axis description cs:-0.07,.5)}, anchor=south},
        grid=major,
        legend style= {at={(0.99,0.55)}, anchor = east},
        scatter/classes={
            0={blue}, 2={green}, 3={orange}, 6={teal},  7={red}
        }
    ]
    
    \pgfplotstableread[col sep=comma]{data/output_reconstruction_transformed.csv}\datatable

    \foreach \vehicle in {0,2,3,6,7} { 
        \addplot[scatter, only marks, scatter src=explicit symbolic, mark size=1pt]
        table [
            x=X_Position,
            y=Y_Position,
            meta=Vehicle_ID,
            row predicate={\thisrow{Vehicle_ID}==\vehicle}
        ] {\datatable};
        
        \addlegendentryexpanded[font=\scriptsize]{Vehicle \vehicle}
        }
    \end{axis}

\end{tikzpicture}
    
\vspace{-8pt}
\caption{Generated vehicle trajectories for the representative scenario of the codebook entry \mbox{$q=167$}. }
    \label{fig:sub1}

\end{subfigure}
    
    \caption{(a) Visualized vehicle trajectories of multiple driving scenarios assigned to the codebook entry $q = 167$ of $Q=253$ and (b) the corresponding generated vehicle trajectories~$\bm{\hat{\xi}}$ for the representative scenario of \mbox{$q=167$}.}
    \label{fig:traj_vis}
    \vspace{-15pt}
\end{figure}

Hence, the architecture also should be capable of predicting the future behavior class more accurately with an increasing number of codebook entries.
To investigate this, confusion matrices are utilized to evaluate the accuracy of the predicted classes of $\mathcal{D}_{\text{test}}$ for $Q_1$,~$Q_2$, and~$Q_3$. 
These matrices, shown in Fig. \ref{fig:confusion_matrices}, illustrate the true versus predicted classifications for the three predefined classes.
All three models exhibit superior performance compared to other class prediction methods from \cite{Neumeier2021} which achieve an accuracy average of 0.76.
However, no improvement in performance is observed with an increasing number of traffic scenario categories.
A performance increase would have been expected with a higher number of codebook entries.
The underlying assumption is that the scenarios within a single scenario category exhibit greater similarity, as evidenced by the lower reconstruction loss.
Since the scenarios are more similar to one another, they are also expected to result in more consistent scenario outcomes.
For instance, in all cases, the target vehicle consistently maintains its lane, since all other lanes are occupied.
The results, however, indicate that this metric could lack sufficient precision.
As a result, this metric alone is inadequate for determining an optimal number of codebook entries.
Finally, the clustering and reconstruction steps are visualized  to provide a comprehensive illustration of the process.
Fig. \ref{fig:traj_vis} shows the traffic scenarios assigned to a specific codebook entry (a) and the reconstructed representative generated by the architecture for exactly that codebook entry (b).
It can be observed that the key features of the input scenarios are also present in the reconstructed scenario.
The patterns are represented in the reconstruction as point clouds that approximate the average of the input data.
This demonstrates that the architecture is effectively capable of clustering and reconstructing traffic scenarios while preserving their essential characteristics.

\subsection{Dataset Completeness} 
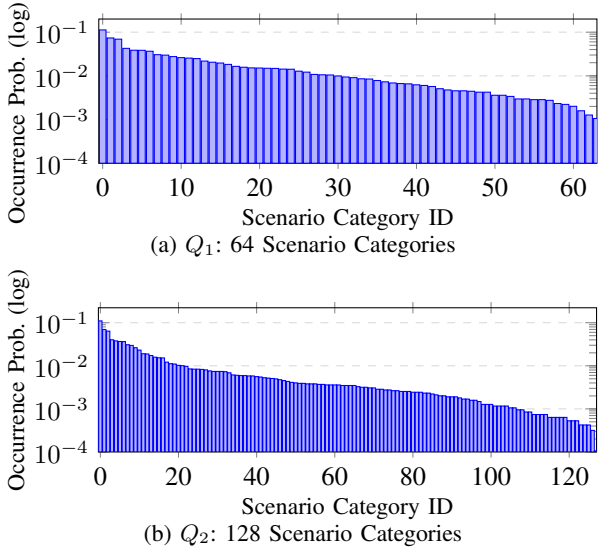
\begin{figure}[t]
    \centering
    \begin{subfigure}[b]{\linewidth}
        \centering
        \begin{tikzpicture}
            \begin{semilogyaxis}[
                log origin=infty,
                ybar,
                ymajorgrids,
                grid style={dashed,gray!30},
                bar width=2.5pt,
                height=3.5cm,
                xtick align=center,
                width=0.95\linewidth,
                xlabel={\small Scenario Category ID},
                ylabel={\small Occurrence Prob. (log)},
                x label style={at={(axis description cs:0.5,-0.25)}, anchor=north},
                y label style={at={(axis description cs:-0.12,.35)}, anchor=south},
                xtick={0,10,20,30,40,50,60},
                xticklabels={0,10,20,30,40,50,60},
                ymin=1e-4, 
                ymax=0.2, 
                xmin=-0.5,
                xmax=63,
                table/col sep=comma
            ]
            \addplot+ table[x=One-Hot-Encoding, y=Probability] {data/CCP_64.csv};
            \end{semilogyaxis}
        \end{tikzpicture}
        \vspace{-8pt}
        \caption{$Q_1$: 64 Scenario Categories}

    \end{subfigure}

    \begin{subfigure}[b]{\linewidth}
        \centering
        \begin{tikzpicture}
            \begin{semilogyaxis}[
                log origin=infty,
                ybar,
                ymode=log,
                ymin=0.00010,
                ymajorgrids,
                grid style={dashed,gray!30},
                bar width=1.25pt,
                height=3.5cm,
                xtick align=center,
                width=0.95\linewidth,
                xlabel={\small Scenario Category ID},
                ylabel={\small Occurrence Prob. (log)},
                x label style={at={(axis description cs:0.5,-0.25)}, anchor=north},
                y label style={at={(axis description cs:-0.12,.5)}, anchor=south},
                xtick={0,20,40,60,80,100,120},
                xticklabels={0,20,40,60,80,100,120},
                xmin=-0.5,
                xmax=127,
                table/col sep=comma
            ]
            \addplot+ table[x=One-Hot-Encoding, y=Probability] {data/CCP_128.csv};
            \end{semilogyaxis}
        \end{tikzpicture}
        \vspace{-8pt}
        \caption{$Q_2$: 128 Scenario Categories}
        \label{fig:model_10}
    \end{subfigure}

    \caption{Histograms showing the scenario category probabilities for (a) $Q_1 =64$ and (b) $Q_2 = 128$.}
    \vspace{-1em}
    \label{fig:Scenario_cat_prob}
\end{figure}

For each of the three scenario category datasets \mbox{$Q_1, Q_2$ and $Q_3$}, the occurrence probability of the respective traffic scenario categories is calculated using \mbox{Equation (\ref{eq:p_q})}.
The results for $Q_1$ with 64 and $Q_2$ with 128 traffic scenario categories are depicted in Fig. \ref{fig:Scenario_cat_prob}, sorted in descending order of occurrence probability.
To compute the required number of simulations $sim$, $R =1,000$ is set in accordance with \cite{Hauer2019}.
The remaining parameters are also defined in accordance, with a standard error of $e = 0.01$ and a confidence level $c_{0.95/2}^2 = 1.96$.
This calculation resulted in an exceptionally high number of required
simulations $sim$, theoretically reaching into the trillions.
To address this limitation, the number of simulations was set to \mbox{$I = 100{,}000$}.
This value was selected as a practical compromise to provide a meaningful estimate of the required sample size while accounting for the existing computational constraints.
To ensure transparency, it must be stated, that fully precise results would require conducting the simulations with the calculated number of $sim$.
As a result, the findings presented here should be regarded as approximations and interpreted within the context of the reduced experiment size.
The experiments are conducted with the confidence level \mbox{$\tau = 0.95$}, along with the values \mbox{$p_{\text{new},1} = 0.001$}, \mbox{$p_{\text{new},2} = 0.0001$}, and \mbox{$p_{\text{new},3} = 0.00001$}.
\begin{table}[t]
    \vspace{1em}
    \centering
    \caption{Calculated number of necessary samples $S_{\text{min}}$ and the standard deviation $\sigma$ showing the influence of $p_{\text{new}}$.}
    \begin{tabular}{ccccc}\\

        \multicolumn{2}{c|}{Dataset} & \multirow{2}{*}{\textit{p}\textsubscript{\text{new}}} & \multicolumn{2}{|c}{$\tau = 0.95$}  \\ \cline{1-2} \cline{4-5}

        \multicolumn{1}{c|}{Source} & \multicolumn{1}{c|}{Categories} & & \multicolumn{1}{|c|}{$S_{\text{min}}$} & \multicolumn{1}{c}{$\sigma$} \\ \toprule \toprule

            Art. dataset \cite{Hauer2019} & 15 & 0.001 & 2,991 & 18.72  \\
            Art. dataset \cite{Hauer2019} & 45 & 0.001 & 3,001 & 21.60 \\ \cline{1-5}
            \ac{CVQ-VAE} & 64 & 0.001 & 3,853 & 710.91  \\
             \ac{CVQ-VAE} & 128 & 0.001 & 29,986 & 5,552.38 \\
             \ac{CVQ-VAE} & 253 & 0.001 & 35,773 & 6,178.96  \\ \hline \hline
            
            Art. dataset \cite{Hauer2019} & 15 & 0.0001 & 29,996 & 165.81 \\
            Art. dataset \cite{Hauer2019} & 45 & 0.0001 & 30,312 & 226.41 \\ \cline{1-5}
             \ac{CVQ-VAE} & 64 & 0.0001 & 29,544 & 6,866.80  \\
            \ac{CVQ-VAE} & 128 & 0.0001 & 35,549 & 7,462.74  \\
             \ac{CVQ-VAE} & 253 & 0.0001 & 39,144 & 6,844.63  \\ \hline \hline
            
            Art. dataset \cite{Hauer2019} & 15 & 0.00001 & 332,544 & 2,111.74   \\
            Art. dataset \cite{Hauer2019} & 45 & 0.00001 & 333,595 & 2,436.66  \\ \cline{1-5}
             \ac{CVQ-VAE} & 64 & 0.00001 & 296,881 & 79,693.96  \\
             \ac{CVQ-VAE} & 128 & 0.00001 & 301,999 & 70,209.30  \\
             \ac{CVQ-VAE} & 253 & 0.00001 & 296,292 & 67,824.61  \\

\end{tabular}
    \vspace{-2em}
    \label{tab:CCP_table}
\end{table}
The results are compared to~\cite{Hauer2019} in \mbox{Table \ref{tab:CCP_table}}.
Firstly, it can be observed that the standard derivation $\sigma$ is higher for the experiments conducted in this study.
This can be attributed, on the one hand, to the considerably lower number of conducted experiments.
On the other hand, it is influenced by the differing dataset structure compared to~\cite{Hauer2019}.
Unlike the artificial datasets used in \cite{Hauer2019}, the datasets in this study include a substantially higher number of scenario categories.
Many of these exhibit relatively low occurrence probabilities.
The comparison of \mbox{$Q_1 = 64$} and \mbox{$Q_3 = 253$} for \mbox{$p_{\text{new},1} = 0.001$} shows a clear increase for $S_{\text{min}}$.
This demonstrates that increasing $Q$ leads to the formation of more scenario categories with lower occurrence probabilities.
As a consequence, more data is required to consider the scenario category datasets complete.
However, this observation contradicts the findings of \cite{Hauer2019}, which suggested that the results are independent of the number of scenario categories.
Since a certain level of confidence is typically required, $p_{\text{new}}$ should always be chosen smaller than the occurrence probability of the already known scenario categories.
This indicates that $Q$ also has a direct impact on $p_{\text{new}}$.
\mbox{Table \ref{tab:CCP_table}} shows that increasing~$p_{\text{new}}$ leads to an increased~$S_{\text{min}}$.
This implies a strong interest in selecting $Q$ as low as possible to minimize the amount of data required to state a scenario category dataset complete.
However, setting ~$Q$ too low introduces the risk to group dissimilar scenarios into the same category.
Therefore, it is essential to gain a detailed understanding of the structure of the scenarios within the dataset to determine the optimal~$Q$.
A central aspect of this process is the development of appropriate metrics to identify the optimal number of scenario categories.
Such metrics are crucial for balancing the trade-off between clustering granularity and data requirements, thereby ensuring an effective evaluation of scenario category completeness.

\section{Conclusions}

This study advances the field of \ac{ADS} by proposing a novel pipeline for traffic scenario clustering and evaluating scenario category completeness.
The implementation and optimization of the \ac{CVQ-VAE} for traffic scenario clustering enabled efficient scenario category generation.
An improved codebook utilization is achieved compared to state-of-the-art approaches. 
This enhancement led to improved performance across the considered metrics, underscoring the effectiveness of the proposed approach.
Furthermore, the generated traffic scenario categories served as a foundation for analyzing dataset completeness.
This revealed a direct relationship between the number of categories and dataset completeness. 
However, despite these achievements, the absence of a reliable metric to determine the optimal number of traffic scenario categories remains a limitation and presents an opportunity for future research.
While this work served as a proof of concept for the proposed pipeline tested on a single dataset, future efforts will evaluate its efficiency across multiple datasets.

\bibliographystyle{IEEEtran}
\bibliography{Literature}

\end{document}